\documentclass[pdflatex,sn-mathphys-num]{sn-jnl}

\usepackage{graphicx}
\usepackage{multirow}
\usepackage{amsmath,amssymb,amsfonts}
\usepackage[title]{appendix}
\usepackage{xcolor}
\usepackage{textcomp}
\usepackage{booktabs}
\usepackage{hyperref}
\usepackage{url}

\renewcommand{\orcidlogo}{\textcolor[HTML]{A6CE39}{\textsuperscript{\textsf{ORCID}}}}
\renewcommand{\orcid}[1]{\href{#1}{\orcidlogo}}

\begin{document}

%%=============================================================%%
%% Title
%%=============================================================%%

\title[MetaboT: LLM-based Multi-Agent Framework for Metabolomics]{MetaboT: An LLM-based Multi-Agent Framework for Interactive Analysis of Mass Spectrometry Metabolomics Knowledge Graphs}

%%=============================================================%%
%% Authors and affiliations
%%=============================================================%%

\author[1,2,3]{\fnm{Madina} \sur{Bekbergenova}}
\author[1,2]{\fnm{Lucas} \sur{Pradi}\orcid{https://orcid.org/0009-0003-8641-6268}}
\author[1,2]{\fnm{Benjamin} \sur{Navet}}
\author[1,2,4,5,6]{\fnm{Emma} \sur{Tysinger}\orcid{https://orcid.org/0000-0002-3958-8097}}
\author[2,6]{\fnm{Franck} \sur{Michel}\orcid{https://orcid.org/0000-0001-9064-0463}}
\author[1,2]{\fnm{Matthieu} \sur{Feraud}\orcid{https://orcid.org/0009-0001-5496-6690}}
\author[1,2,6]{\fnm{Yousouf} \sur{Taghzouti}\orcid{https://orcid.org/0000-0003-4509-9537}}
\author[7]{\fnm{Yan Zhou} \sur{Chen}}
\author[8,9]{\fnm{Olivier} \sur{Kirchhoffer}\orcid{https://orcid.org/0000-0001-7567-1050}}
\author[10]{\fnm{Florence} \sur{Mehl}\orcid{https://orcid.org/0000-0002-9619-1707}}
\author[1,2]{\fnm{Martin} \sur{Legrand}}
\author[1,2]{\fnm{Tao} \sur{Jiang}\orcid{https://orcid.org/0000-0002-5293-3916}}
\author[10]{\fnm{Marco} \sur{Pagni}\orcid{https://orcid.org/0000-0001-9292-9463}}
\author[7,11]{\fnm{Soha} \sur{Hassoun}\orcid{https://orcid.org/0000-0001-9477-2199}}
\author[8,9]{\fnm{Jean-Luc} \sur{Wolfender}\orcid{https://orcid.org/0000-0002-0125-952X}}
\author*[3]{\fnm{Wout} \sur{Bittremieux}\orcid{https://orcid.org/0000-0002-3105-1359}}\email{wout.bittremieux@uantwerpen.be}
\author*[2,6]{\fnm{Fabien} \sur{Gandon}\orcid{https://orcid.org/0000-0003-0543-1232}}\email{fabien.gandon@inria.fr}
\author*[1,2]{\fnm{Louis-F{\'e}lix} \sur{Nothias}\orcid{https://orcid.org/0000-0001-6711-6719}}\email{louis-felix.nothias@cnrs.fr}

\affil[1]{\orgname{Universit{\'e} C\^ote d'Azur, CNRS, ICN}, \orgaddress{\city{Nice}, \country{France}}}
\affil[2]{\orgname{Interdisciplinary Institute for Artificial Intelligence (3iA) C\^ote d'Azur}, \orgaddress{\city{Sophia-Antipolis}, \country{France}}}
\affil[3]{\orgdiv{Department of Computer Science}, \orgname{University of Antwerp}, \orgaddress{\city{Antwerp}, \country{Belgium}}}
\affil[4]{\orgname{Massachusetts Institute of Technology}, \orgaddress{\city{Cambridge}, \state{MA}, \country{USA}}}
\affil[5]{\orgname{:probabl.}, \orgaddress{\city{Paris}, \country{France}}}
\affil[6]{\orgname{Inria, Universit{\'e} C\^ote d'Azur, CNRS, I3S}, \orgaddress{\country{France}}}
\affil[7]{\orgdiv{Department of Computer Science}, \orgname{Tufts University}, \orgaddress{\city{Medford}, \state{MA 02155}, \country{USA}}}
\affil[8]{\orgname{Institute of Pharmaceutical Sciences of Western Switzerland, University of Geneva, Centre M{\'e}dical Universitaire}, \orgaddress{\city{Geneva}, \country{Switzerland}}}
\affil[9]{\orgname{School of Pharmaceutical Sciences, University of Geneva, Centre M{\'e}dical Universitaire}, \orgaddress{\city{Geneva}, \country{Switzerland}}}
\affil[10]{\orgname{Swiss Institute of Bioinformatics (SIB)}, \orgaddress{\city{Lausanne}, \country{Switzerland}}}
\affil[11]{\orgdiv{Department of Chemical and Biological Engineering}, \orgname{Tufts University}, \orgaddress{\city{Medford}, \state{MA 02155}, \country{USA}}}

%%=============================================================%%
%% Abstract
%%=============================================================%%

\abstract{Mass spectrometry-based metabolomics generates complex, high-dimensional data that holds vast potential for biological discovery but remains difficult to integrate and interpret. Knowledge graphs (KGs) unify this heterogeneous information by representing spectra, annotations, taxa, chemical classes, and biological activities as a single interoperable network; however, their practical use is limited by the steep learning curve of corresponding specialized representation and query languages. Here we introduce MetaboT, an open-source multi-agent Large Language Model (LLM) framework that translates natural-language questions into executable SPARQL queries over metabolomics knowledge graphs. MetaboT mitigates the hallucination and schema-compliance limitations of single-model approaches through a modular architecture in which specialised agents handle scope validation, entity resolution against authoritative resources, schema-aware query generation, iterative refinement, and result interpretation. We validated MetaboT on the Experimental Natural Products Knowledge Graph (ENPKG), using an expert-authored benchmark of natural-language questions paired with reference SPARQL queries, and demonstrate its ability to answer complex questions about plant--metabolite relationships and biological activities. MetaboT lowers the technical barrier for metabolomics researchers and enables semantic data mining without specialised programming expertise.}


\maketitle

%%=============================================================%%
%% Scientific Contribution (JCim Software article requirement)
%%=============================================================%%

\section*{Scientific Contribution}

MetaboT introduces a multi-agent LLM architecture that grounds each step of natural-language-to-SPARQL translation in authoritative identifier resolution and schema-aware validation, substantially reducing the hallucinations observed with single-model baselines. The framework couples orchestrated agents with domain-specific tools for chemical, taxonomic, and biological target resolution, and with an iterative refinement loop that distinguishes query-construction errors from genuine data absence. Released as open-source software with an archived evaluated version, a reusable benchmark, and a public web demonstrator, MetaboT provides a reproducible reference implementation for knowledge-graph-driven exploration of mass spectrometry metabolomics data.

%%=============================================================%%
%% Introduction
%%=============================================================%%

\section{Introduction}\label{sec:intro}

Contemporary mass spectrometry-based metabolomics generates unprecedented volumes of spectral data, with a single untargeted experiment routinely producing thousands of fragmentation spectra~\cite{Guo:2022:metabolomics-bigdata}. While these raw data contain vast potential for discovery, they remain uninformative without extensive computational processing~\cite{Guo:2022:metabolomics-bigdata}. Over the past decade, the community has developed a rich ecosystem of open-source frameworks that streamline spectral data processing and annotation, including MZmine~\cite{Schmid:2023:MZmine3}, SIRIUS~\cite{Duhrkop:2021:CANOPUS}, GNPS~\cite{Wang:2016:GNPS}, and OpenMS~\cite{Rost:2016:OpenMS}. Yet the complexity and volume of metabolomics data continue to demand advanced expertise in statistical analysis, programming, data engineering, and visualisation, ultimately constraining researchers' ability to extract deeper biological insights~\cite{Alseekh:2021:metabolomics-guide}.

Over the past decade, molecular networking has democratised mass spectra exploration by enabling researchers from diverse backgrounds to intuitively map structural relationships among metabolites through graph-based comparison of fragmentation spectra~\cite{Wang:2016:GNPS}. More recently, knowledge graphs have emerged as the next frontier in the field, semantically integrating metabolites, pathways, taxa, and biological activities into an interoperable framework~\cite{Ebbels:2023:computational-metabolomics,Meijer:2025:AI-natural-products,Gaudry:2024:ENPKG}. These graphs rely on domain-specific ontologies that provide standardised vocabularies, including concepts and relation hierarchies, essential for consistent data integration across different experimental platforms. Despite their potential, knowledge graphs and their underlying ontologies remain underutilised, because of the steep learning curve associated with representation languages such as RDF and query languages such as SPARQL, as well as the complexity of mastering domain-specific ontological frameworks. These technical barriers require in-depth understanding of the data graph model, its query syntax, and the underlying schema that defines its structure, constraints, and semantics---significantly limiting accessibility for many researchers~\cite{Rony:2022:SGPT,Perevalov:2024:understanding-SPARQL}.

Recent advances in generative AI using Large Language Models (LLMs) have shown promising results in automating SPARQL query generation~\cite{Rony:2022:SGPT,Meyer:2024:SPARQL-LLM,Emonet:2024:ExpasyGPT,Sima:2023:AI-chatbots-KG} and in enabling knowledge-graph construction through tools such as BioCypher~\cite{Lobentanzer:2023:BioCypher}. However, LLMs still suffer from critical limitations. In particular, they are prone to hallucinations, generating incorrect identifiers or properties and thereby corrupting query construction~\cite{Sima:2023:AI-chatbots-KG}. Their performance is also heavily dependent on access to detailed schema information and precise entity identifiers, which has so far limited their standalone applicability in complex metabolomics contexts. Supplementing LLMs with tools such as retrieval-augmented generation (RAG) pipelines, schema parsers, specialised entity-resolution modules, or integrated frameworks such as BioChatter and ExpasyGPT~\cite{Emonet:2024:ExpasyGPT,Lobentanzer:2025:BioChatter} can reduce dependency on schema memorisation, yet these approaches still demand complex, schema-specific integration and may hallucinate identifiers because of limited orchestration. Emerging evidence shows that SPARQL generation improves markedly with chain-of-thought prompting~\cite{Zahera:2024:SPARQL-CoT} and role-based specialised-agent orchestration~\cite{Zong:2024:Triad}.

\begin{figure}[ht]
\centering
\includegraphics[width=\textwidth]{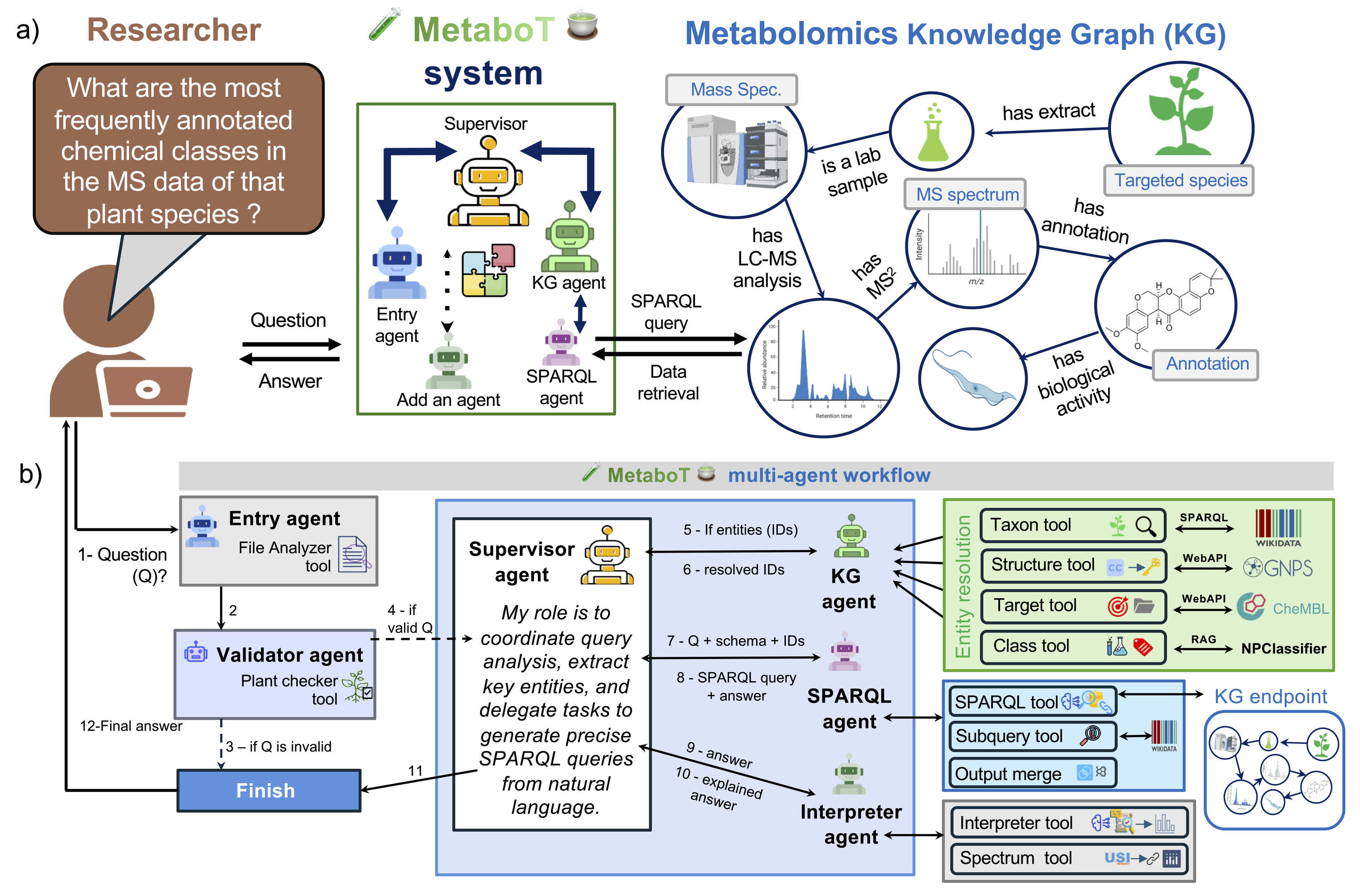}
\caption{MetaboT overview. \textbf{(a)} A user's natural-language question is translated into a SPARQL query and executed against the mass spectrometry knowledge graph, returning the final answer. \textbf{(b)} Detailed schematic of the multi-agent architecture, highlighting specialised agents and tools for entity resolution, query validation, and result interpretation. The ``SPARQL Agent'' in the figure corresponds to the \textit{SPARQL Query Runner Agent} described in the main text.}\label{fig:overview}
\end{figure}

To overcome these hurdles---and to harness the power of chain-of-thought prompting and role-based agent orchestration---we developed MetaboT, a multi-agent LLM framework that decomposes complex metabolomics knowledge-graph queries into discrete subtasks, routes each to a specialist agent, and thereby delivers precise entity resolution, schema-aware SPARQL generation, and markedly fewer hallucinations than single-LLM baselines. MetaboT enables researchers to query a metabolomics knowledge graph through an intuitive conversational interface (Fig.~\ref{fig:overview}a). It resolves entity identifiers through orchestrated resolution steps to prevent hallucinations, leverages knowledge-graph schemas, and translates natural-language questions into SPARQL queries adapted to the specific schema of the targeted graph. MetaboT is released as open-source software under the Apache 2.0 license, with an archived evaluated version, a reusable benchmark, and a public web demonstrator, and constitutes the founding proof-of-concept prototype of the MetaboLinkAI programme (\url{https://www.metabolinkai.net}) for open AI-assisted metabolomics.

%%=============================================================%%
%% Implementation
%%=============================================================%%

\section{Implementation}\label{sec:implementation}

MetaboT is implemented as a modular, multi-agent pipeline designed to manage the complexities of natural-language-to-SPARQL translation (Fig.~\ref{fig:overview}b). The system orchestrates tasks through a hierarchy of specialised agents and supporting tools, built on top of LangChain~\cite{LangChain:2024} (v0.3.27) and LangGraph~\cite{LangGraph:2024} (v0.3.34). LangChain provides a flexible framework for developing LLM-based applications, with standard components for LLM calls, prompts, chains, and tool-using agents. LangGraph structures the workflow as a stateful directed graph, where nodes represent discrete computational steps and edges represent message passing between agents. This stateful graph enables dynamic state evolution as each agent completes its task, allowing for adaptive interactions throughout the query-processing pipeline. Formally, each agent in the workflow is represented as a tuple $A_i = (L_i, P_i, T_i)$~\cite{Topsakal:2023:LLM-apps}, where $L_i$ is the language model guiding decision-making, $P_i$ is the prompt customised for the agent's role, and $T_i$ is the set of tools the agent can invoke. The system was initially developed and evaluated using GPT-4o and GPT-4o-mini. Because the architecture is model-agnostic, the LLM backend is configurable; the current implementation updates this default to GPT-5.5 while maintaining support for other models.

\subsection{Schema and identifier provision}\label{subsec:schema}

To accurately generate SPARQL queries from natural-language inputs, the LLM must be supplied with the specific ontology schema and Internationalized Resource Identifiers (IRIs) pertinent to the target knowledge graph: the schema defines the relationships between nodes~\cite{Hogan:2021:knowledge-graphs} and IRIs uniquely identify the entities and properties used in queries~\cite{BernersLee:2005:URI}. LLMs are trained on broad datasets that may not include these specific IRIs~\cite{Meyer:2024:SPARQL-LLM,Sima:2023:AI-chatbots-KG}, and even when relevant IRIs are present in the training data, LLMs are not guaranteed to retrieve them correctly or to incorporate them appropriately into SPARQL queries, limiting their effectiveness for precise identifier-based tasks. Consequently, these IRIs must be supplied externally to the LLM alongside the ontology schema to ensure correct SPARQL construction. Prior work by Meyer et al.~\cite{Meyer:2024:SPARQL-LLM} and Avila et al.~\cite{Avila:2024:NL-SPARQL} confirms that providing LLMs with explicit schema information and IRIs significantly enhances the accuracy and reliability of SPARQL query generation. MetaboT therefore integrates dedicated tools and agents that resolve entities against authoritative resources and supply IRIs together with the relevant Turtle-serialised schema before query generation.

\subsection{Workflow and agent orchestration}\label{subsec:workflow}

MetaboT processes user inputs through a flexible pipeline that manages context, entity resolution, query generation, and iterative refinement (Fig.~\ref{fig:overview}b). The execution lifecycle proceeds as follows:

\begin{enumerate}

\item \textbf{Context management and validation.} When a user submits a natural-language question (e.g., \textit{``What bioactive compounds are found in Tabernaemontana coffeoides?''}), the \textit{Entry Agent} first determines whether the question is new or a follow-up to previous interactions. New questions are forwarded to the \textit{Validator Agent}, a single LLM prompted with relevant details of the knowledge-graph schema, which checks the question's scope to ensure relevance (e.g., confirming that bioactive compounds and plant species are valid entity types in the schema). For plant-specific questions, the \textit{PlantDatabaseChecker} tool confirms the presence of the plant in a curated database (e.g., verifying that \textit{Tabernaemontana coffeoides} exists in the plant database). Upon successful validation, the question is passed to the \textit{Supervisor Agent}.

\item \textbf{Supervisor delegation and entity resolution.} The \textit{Supervisor Agent} acts as the central coordinator for verified queries and routes them to appropriate agents based on content analysis. In our demonstration we use the Experimental Natural Products Knowledge Graph (ENPKG)~\cite{Gaudry:2024:ENPKG}, a comprehensive semantic graph that integrates mass spectrometry data with taxonomic information, chemical structures, and biological activities of natural products from plant sources. When entity resolution is required, the \textit{Supervisor Agent} delegates to the \textit{KG Agent}, which extracts entities mentioned in the user's question and resolves them to their IRIs via specialised tools:
  \begin{itemize}
    \item When \textit{``Tabernaemontana coffeoides''} is mentioned, the \textit{KG Agent} calls the \textit{TaxonResolver} to retrieve its Wikidata IRI.
    \item For chemical classes such as \textit{``flavonoids''}, the \textit{ChemicalResolver} tool retrieves the appropriate NPClassifier (NPCClass) IRI.
    \item When a SMILES string is provided (e.g., \textit{``C1=CC=C2C(=C1)C=CN2''}), the \textit{SMILESResolver} converts it to a standardised InChIKey.
    \item When a biological target such as \textit{``acetylcholinesterase''} is referenced, the \textit{TargetResolver} obtains its ChEMBL target IRI.
  \end{itemize}
  Once entities are resolved, the \textit{KG Agent} verifies the returned matches against contextual information and returns precise identifiers to the \textit{Supervisor}.

\item \textbf{Query generation and execution.} With entities resolved and schema details collated, the \textit{Supervisor Agent} routes the request to the \textit{SPARQL Query Runner Agent}. This agent contextualises the user's question with the aligned identifiers and a Turtle-formatted schema to build a structured prompt for the \textit{GraphSparqlQAChain} tool, which generates a SPARQL SELECT query, executes it on the KG endpoint, and retrieves the answer.

\item \textbf{Interpretation.} If the SPARQL output is large or the user explicitly requests a visualisation (e.g., \textit{``Plot the distribution''}), the \textit{Interpreter Agent} is invoked, which delegates the task to the \textit{Interpreter tool}. This tool dynamically generates and executes Python code within a secure, locally hosted subprocess sandbox to produce textual summaries and, on demand, Plotly-based visualisations (e.g., bar charts, scatter plots) as JSON specifications. These visualisations are rendered automatically in the Streamlit web application; when running locally, the JSON specifications are returned without automatic rendering.

\end{enumerate}

To manage computational costs and context-window limits, the system monitors token usage and systematically generates a CSV file containing the complete results. If the combined tokens of the query context and KG output exceed 6,000, the system suppresses the direct text output in the interface and instead provides the user solely with the file path to the generated CSV.

\subsection{Agent framework}\label{subsec:agents}

MetaboT implements six primary agents, each with a distinct role in processing user queries.

\textbf{Entry Agent.} \textit{Purpose:} gateway to the MetaboT system; determines whether a query is new or related to prior interactions. \textit{Key functions:} processes user-provided files via the \textit{FileAnalyzer} tool; forwards new queries to the \textit{Validator Agent}; retrieves context for follow-up questions from the conversation history. \textit{Processing approach:} classifies queries as \textit{new knowledge questions} (requiring new database information) or \textit{help me understand questions} (seeking clarification on previous answers), preventing redundant processing and maintaining conversation context.

\textbf{Validator Agent.} \textit{Purpose:} ensures that user questions are appropriate for querying the knowledge graph before further processing. \textit{Key functions:} confirms query validity and alignment with the schema and available data; uses the \textit{PlantDatabaseChecker} tool to verify the presence of mentioned plant or taxon names; routes valid questions to the \textit{Supervisor Agent} and returns specific feedback for invalid ones. \textit{Processing approach:} employs in-context learning~\cite{dong-etal-2024-survey} and few-shot examples by embedding a compact form of the knowledge-graph schema directly in its prompt, enabling schema compliance through comparison with valid and invalid example questions.

\textbf{Supervisor Agent.} \textit{Purpose:} central coordinator; routes user queries to appropriate agents based on content analysis. \textit{Key functions:} identifies key entities (taxon names, chemical names, biological targets, etc.); delegates entity resolution to the \textit{KG Agent} when needed; passes processed outputs to the \textit{SPARQL Query Runner Agent}. \textit{Processing approach:} follows explicit guidelines for task delegation, entity-specific processing, and multi-step query workflows, with worked examples covering proper task routing and error handling.

\textbf{KG Agent.} \textit{Purpose:} resolves entities mentioned in user queries to provide precise identifiers for knowledge-graph elements. \textit{Key functions:} retrieves standardised identifiers through specialised resolver tools---\textit{ChemicalResolver} (RAG over a reference file), \textit{SMILESResolver} (GNPS API), \textit{TargetResolver} (ChEMBL API), and \textit{TaxonResolver} (SPARQL over the Wikidata endpoint). \textit{Processing approach:} uses external APIs, RAG on local files, or SPARQL queries to resolve each entity against authoritative sources rather than relying on the LLM's parametric knowledge, substantially reducing hallucinations.

\textbf{SPARQL Query Runner Agent.} \textit{Purpose:} prepares inputs for SPARQL generation and coordinates query execution. \textit{Key functions:} if a question requires Wikidata comparison, extracts the Wikidata-related part (using \textit{WikidataStructureSearch}) and sends only the non-Wikidata component with resolved entities to the \textit{GraphSparqlQAChain} tool; otherwise, selects \textit{GraphSparqlQAChain} and provides the complete question with resolved entities; ensures all required information (user question, resolved entities, schema details) is properly structured; manages query execution and result formatting; when applicable, calls \textit{OutputMerger} to combine \textit{GraphSparqlQAChain} and \textit{WikidataStructureSearch} outputs and returns the merged result to the \textit{Supervisor}. \textit{Processing approach:} emphasises structured input preparation rather than direct query generation, ensuring precise SPARQL construction through comprehensive context provision.

\textbf{Interpreter Agent.} \textit{Purpose:} bridges the gap between raw query results and user-friendly insights. \textit{Key functions:} analyses outputs from the \textit{SPARQL Query Runner Agent} or user-submitted files; produces clear summaries of query results; generates visualisations (bar charts, diagrams, mass-spectrum plots) when required. \textit{Processing approach:} invokes the \textit{Interpreter} tool with structured inputs; follows detailed guidelines for presenting data in an accessible, query-appropriate format.

\subsection{Tool integration}\label{subsec:tools}

MetaboT employs eleven specialised tools that support the agents and integrate external resources.

\begin{enumerate}

\item \textbf{PlantDatabaseChecker.} Verifies the presence of plant names within a pre-existing database to filter queries based on data availability. \textit{Input:} plant name string. \textit{Process:} loads a CSV file containing plant names considered in the ENPKG analysis and searches for exact matches. \textit{Output:} confirmation of the plant's presence or absence, with error handling for file access issues.

\item \textbf{ChemicalResolver.} Resolves chemical names to standardised NPClassifier class identifiers (NPCClass URIs). \textit{Input:} chemical name extracted from the user query. \textit{Process:} matches chemical names to NPCClass URIs using a FAISS~\cite{Johnson:2021:FAISS} vector store with OpenAI embeddings~\cite{OpenAI:2024:embeddings}, built from a precompiled CSV file of NPCClass data extracted from the knowledge graph. \textit{Output:} NPCClass URI with type information.

\item \textbf{SMILESResolver.} Converts SMILES~\cite{Weininger:1988:SMILES} molecular representations to standardised InChIKeys~\cite{Heller:2013:InChI} for chemical interoperability. \textit{Input:} SMILES string. \textit{Process:} sends the SMILES to the GNPS API to retrieve the corresponding InChIKey. \textit{Output:} standardised InChIKey with process logging. \textit{Error handling:} robust logging for API request issues and exception handling.

\item \textbf{TargetResolver.} Maps biological target names to standardised ChEMBL target IRIs~\cite{Zdrazil:2024:ChEMBL}. \textit{Input:} biological target name. \textit{Process:} queries the ChEMBL API to locate matches based on target names, parsing XML responses to extract and format ChEMBL target IDs as IRIs. \textit{Output:} formatted ChEMBL target IRI. \textit{Error handling:} logs successful and unsuccessful API calls and suggests refinement when no matches are found.

\item \textbf{TaxonResolver.} Resolves taxonomic names to their corresponding Wikidata IRIs. \textit{Input:} taxonomic name. \textit{Process:} constructs and executes SPARQL queries on the Wikidata endpoint~\cite{Waagmeester:2020:Wikidata} to retrieve unique IRIs associated with taxon names. \textit{Output:} Wikidata IRI for the taxonomic entity. \textit{Error handling:} robust handling of unexpected formats and connectivity issues with clear logging.

\item \textbf{FileAnalyzer.} Analyses user-provided files and generates content summaries. \textit{Input:} uploaded files and session identifier. \textit{Process:} initialises session-specific directories, identifies file types, and applies tailored analysis methods---row/column/header extraction via pandas for spreadsheets, spectrum counts for Mascot Generic Format (MGF) files, and line counts for text files. \textit{Output:} structured summaries with file paths, sizes, and content-specific details for downstream processing.

\item \textbf{GraphSparqlQAChain.} Generates and executes SPARQL queries aligned with the knowledge graph's schema. \textit{Input:} user question, schema information in Turtle format, and resolved entity identifiers. \textit{Process---query generation:} uses the LLM with structured prompts combining in-context learning, chain-of-thought reasoning~\cite{Zahera:2024:SPARQL-CoT}, and one-shot examples to create schema-compliant queries. \textit{Process---query refinement:} when initial queries yield no results, identifies schema nodes related to the original query and leverages a FAISS store of similar queries to regenerate a more accurate command. \textit{Process---result processing:} executes queries on the KG endpoint and saves outputs to temporary CSV files. \textit{Error handling:} removes extraneous elements from generated queries, ensures schema compliance, and manages token limits by truncating outputs for the LLM context window when necessary.

\item \textbf{WikidataStructureSearch.} Retrieves compound annotations associated with a taxon (e.g., plant genus) using its Wikidata identifier. \textit{Input:} Wikidata identifier string. \textit{Process:} constructs and executes a SPARQL query on the Wikidata endpoint to retrieve Wikidata IRIs of chemical compounds annotated to species within the genus of the given taxon; results are saved to a temporary CSV file. \textit{Output:} path to a CSV file containing the list of compound Wikidata IRIs, or \texttt{None} if no results are found.

\item \textbf{OutputMerger.} Identifies overlapping Wikidata compound annotations between results from two different SPARQL endpoints. \textit{Input:} two CSV paths containing Wikidata IDs---one from the ENPKG endpoint (\textit{GraphSparqlQAChain}) and one from the Wikidata endpoint (\textit{WikidataStructureSearch}). \textit{Process:} standardises identifier formatting, identifies common IDs, removes duplicates, and saves the final list to a temporary CSV file. \textit{Output:} path to the merged CSV.

\item \textbf{Interpreter.} Contextualises query results through summaries and visualisations. \textit{Input:} SPARQL query results, file paths, and the original user question. \textit{Process:} parses agent inputs to extract questions, file paths, and contextual information; delegates execution to the \textit{Interpreter tool}, which dynamically generates and runs Python code within a secure, custom-built local subprocess sandbox (restricted to user-specific session directories) to produce interpretations of query results and Plotly~\cite{Plotly:2015} JSON graph specifications on request. \textit{Output:} textual summaries and, when visualisations are requested, JSON files containing Plotly graph code. \textit{Output management:} logs processing steps and provides accessible file paths to users.

\item \textbf{SpectrumPlotter.} Provides a URL that displays a plot of a spectrum based on a given Universal Spectrum Identifier (USI)~\cite{HUPO:2023:USI}. \textit{Input:} a USI value that uniquely identifies the spectrum, or a CSV file path with a column named \texttt{usi}. \textit{Process:} constructs a URL of the form \texttt{https://metabolomics-usi.gnps2.org/dashinterface/?usi1=\{usi\}} that points to the Metabolomics Spectrum Resolver dashboard~\cite{Bittremieux:2020:SpectrumResolver}. \textit{Output:} the constructed URL.

\end{enumerate}

\subsection{SPARQL query generation and execution}\label{subsec:sparql}

The \textit{GraphSparqlQAChain} tool implements a three-stage process for translating natural-language questions into executable SPARQL queries.

\textbf{Initial query generation.} The tool combines the user's question with schema information (in Turtle format) and the resolved entity identifiers to generate a SPARQL SELECT query. For example, a user question about \textit{``bioactive compounds in Tabernaemontana coffeoides''} is combined with the Wikidata IRI for \textit{Tabernaemontana coffeoides} (Q15376858) and the relevant schema properties. The generation prompt enforces content clarity, proper prefix usage, schema alignment, and validation requirements, and combines in-context learning, chain-of-thought prompting~\cite{Zahera:2024:SPARQL-CoT}, and one-shot examples to guide the LLM toward well-structured, syntactically correct queries.

\textbf{Query execution and result analysis.} Generated queries are executed on the knowledge-graph endpoint, with results analysed for completeness and relevance. If a query yields no results or errors, the tool initiates a refinement process.

\textbf{Query refinement.} For unsuccessful queries, the tool identifies schema nodes related to the initial query and retrieves one similar query from a FAISS store of prior queries. Using these insights, it prompts the LLM to regenerate a more accurate, schema-compliant query. This refinement is performed only once, guided by concise schema information, step-by-step reasoning, and validation criteria. An important component of the refinement workflow is the tool's ability to distinguish between errors in query construction and the actual absence of relevant data in the knowledge graph. For example, if a query about \textit{``terpenes in Tabernaemontana coffeoides''} returns no results, the system first assesses whether the query might be improperly constructed and attempts to regenerate it. If the regenerated query returns results, the initial failure is attributed to construction errors. If the regenerated query still fails, this either indicates that the knowledge graph truly lacks data on the topic or that errors persist in the query formulation. The distinction enables targeted refinement strategies that enhance system reliability and user confidence.

\textbf{Result processing.} Successfully executed query results are saved to a temporary CSV file for downstream analysis. Token usage is monitored to ensure compatibility with the LLM's context window; if the combined tokens from the generated SPARQL query, the user's question, and the KG output exceed 6,000 tokens, the full results are not passed to the LLM and users retrieve the complete output from the generated CSV file.

\subsection{Interaction between agents}\label{subsec:interaction}

Agent interaction is orchestrated through a directed graph, with messages passing along edges from one agent to the next (Fig.~\ref{fig:overview}b; Additional file 1, Fig.~S1). The output of each agent becomes the input for subsequent agents, enabling seamless information flow and decision-making throughout the system. For instance, after the \textit{Entry Agent} determines a query's nature, it passes relevant data to the \textit{Validator Agent} via their connecting edge; valid questions proceed to the \textit{Supervisor Agent}, where an initial entity analysis determines whether to invoke the \textit{KG Agent} or proceed directly to SPARQL query generation. The structured directed-graph communication ensures that each agent operates with the most current and relevant information, enhancing efficiency and accuracy. During development and evaluation, agent interactions were monitored and analysed using LangSmith~\cite{LangSmith:2024} (v0.3.45), which we used to systematically evaluate agent performance, track interactions, and identify refinement opportunities.

\subsection{Web application}\label{subsec:webapp}

\begin{figure}[ht]
\centering
\includegraphics[width=\textwidth]{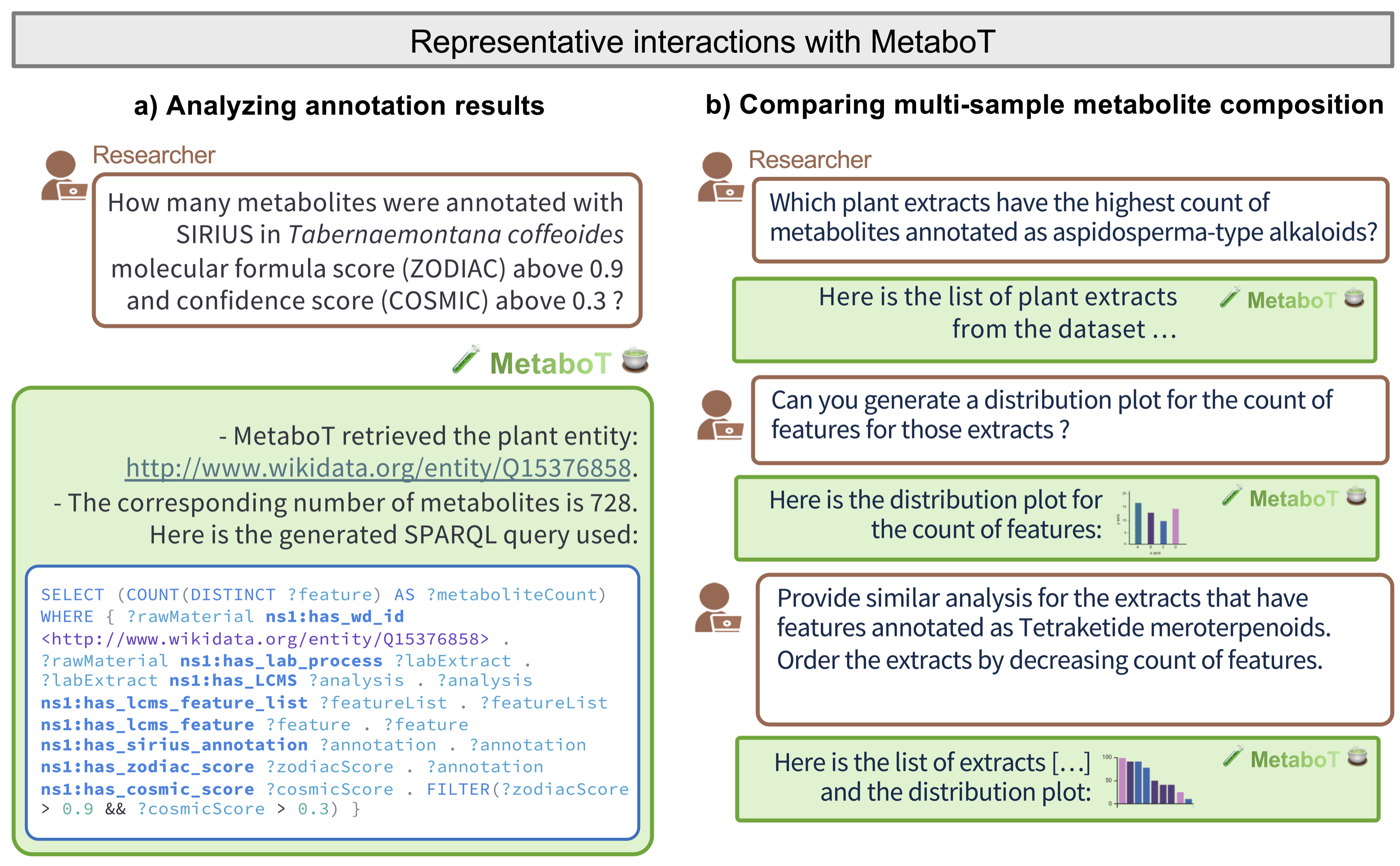}
\caption{Representative interactions with MetaboT. \textbf{(a)} Annotation query. From \textit{Tabernaemontana coffeoides}, MetaboT retrieves 728 metabolites annotated by SIRIUS (ZODIAC $> 0.9$, COSMIC $ > 0.3 $) and displays the generated SPARQL. \textbf{(b)} Cross-sample comparison. MetaboT ranks extracts by counts of aspidosperma alkaloids, plots the distribution, and repeats the analysis for tetraketide meroterpenoids. Green panels are MetaboT responses; brown panels are researcher prompts.}\label{fig:interactions}
\end{figure}

MetaboT is accessible via a web application (\url{https://metabot.holobiomicslab.eu}) designed to prioritise usability for researchers (Fig.~\ref{fig:interactions}). It captures natural-language questions and displays results, while the processing is managed by the backend LLM multi-agent framework. Current features---file uploads, visualisations, and the \textit{SpectrumPlotter} module that integrates Metabolomics USIs~\cite{HUPO:2023:USI} to render spectrum plots~\cite{Bittremieux:2020:SpectrumResolver}---are implemented as proofs of concept that demonstrate the system's extensibility. Outputs are presented as concise summaries, optionally supplemented with plots for visual interpretation of complex datasets. MetaboT's design ensures accessibility for researchers without extensive computer-science expertise while providing robust data-querying capabilities through an interactive query interface, automated SPARQL generation, and built-in error handling.

%%=============================================================%%
%% Results
%%=============================================================%%

\section{Results}\label{sec:results}

\subsection{Validation dataset}\label{subsec:dataset}

To evaluate MetaboT's performance, we constructed a synthetic dataset of 50 natural-language questions paired with manually verified reference SPARQL queries, designed to span multiple ontology classes within ENPKG. The questions were drafted with domain experts, considering the classes and properties present in the schema of the knowledge graph. Each question was submitted to MetaboT to generate an initial SPARQL query, which was manually reviewed and corrected against the ENPKG schema to ensure semantic correctness and full alignment with the intent of the question. Questions were categorised by complexity:

\begin{itemize}
  \item \textbf{Low complexity} (11 questions): simple entity retrieval (e.g., \textit{``List all compounds found in Lovoa trichilioides''}).
  \item \textbf{Medium complexity} (19 questions): multi-step operations with filtering (e.g., \textit{``Find all plant species containing flavonoids with molecular $weight > 300$''}).
  \item \textbf{High complexity} (20 questions): complex relationships and filtering (e.g., \textit{``Which compounds in Acer species have been reported to interact with human kinase targets and show molecular similarity (cosine score) $> 0.7$ to known anti-inflammatory compounds?''}).
\end{itemize}

The complete dataset is available in the project repository (\texttt{app/data/evaluation\_dataset.csv}) and archived on Zenodo (DOI 10.5281/zenodo.19715403).

\subsection{Evaluation procedure}\label{subsec:eval-procedure}

We implemented an automated evaluation process in November 2024 using the LangSmith evaluation workflow~\cite{LangSmith:2024}, with LLM-based evaluators assessing outputs against criteria including correctness, accuracy, and SPARQL query similarity to the reference. To mitigate the known tendency of LLMs to hallucinate during evaluation~\cite{gu2026survey}, we complemented the automated scores with manual review.

A baseline evaluation established whether GPT-4o alone---without the multi-agent system---could produce accurate SPARQL. Questions were submitted to GPT-4o together with the schema in Turtle format, using a prompt analogous to that of the query-generation step of \textit{GraphSparqlQAChain}. Results showed that standalone GPT-4o produced inaccurate queries with property mismatches and incorrect entity identifiers, motivating the structured multi-agent approach.

Building on this baseline, we compared three configurations:

\begin{enumerate}
  \item GPT-4o in single-shot mode without orchestration.
  \item MetaboT using GPT-4o mini as the underlying LLM.
  \item MetaboT using GPT-4o within the full multi-agent framework.
\end{enumerate}

Each configuration processed identical queries under controlled conditions (consistent hardware and network). The evaluation was performed up to the point of SPARQL query generation and did not include the \textit{Interpreter Agent}; the goal was to ensure that all components leading to SPARQL generation functioned correctly.

Across the 50 questions, the refinement chain of \textit{GraphSparqlQAChain} regenerated three queries using formulations identical to the correct reference queries. This refinement was beneficial for only one of the three questions, which was therefore excluded from the overall accuracy computation to avoid crediting the framework with a coincidentally matching reformulation. In the two other cases, the underlying issue was that the \textit{Supervisor Agent} had not invoked the \textit{KG Agent} to retrieve the necessary entities. Accuracy reported below is therefore computed over \textbf{49 questions}, while the full 50-question dataset is preserved and released for future evaluations.

\subsection{Overall and per-complexity accuracy}\label{subsec:accuracy}

MetaboT using GPT-4o within the multi-agent framework achieved 83.67\% overall accuracy on the 49-question evaluation set, compared with 8.16\% for single-shot GPT-4o and 12.24\% for MetaboT using GPT-4o mini (Table~\ref{tab:accuracy}). The gap between the standalone and multi-agent GPT-4o configurations was most pronounced for high-complexity queries, where the multi-agent GPT-4o configuration maintained 78.95\% accuracy while standalone GPT-4o failed completely (0\%). The GPT-4o mini-based MetaboT system performed better on high-complexity questions than on low-complexity ones; this counterintuitive pattern is explained by the fact that low-complexity questions often omit explicit entity mentions, making them harder for the \textit{Validator Agent}---powered by a weaker LLM---to accept.

\begin{table}[ht]
\caption{Accuracy of the three evaluated configurations across complexity levels. Accuracy is reported as the percentage of correctly generated SPARQL queries over 49 evaluation questions (Low: 11; Medium: 19; High: 19). Dominant error types are indicated in parentheses (see Section~\ref{subsec:errors}).}\label{tab:accuracy}
\begin{tabular*}{\textwidth}{@{\extracolsep\fill}lcccc}
\toprule
\textbf{System} & \textbf{Overall (\%)} & \textbf{Low (\%)} & \textbf{Medium (\%)} & \textbf{High (\%)} \\
\midrule
GPT-4o (single-shot)  & 8.16  & 18.18 (T2) & 10.53 (T2)    & 0.00 (T2)    \\
MetaboT -- GPT-4o mini & 12.24 & 0.00 (T1)  & 15.79 (T1--4) & 15.79 (T1--4) \\
MetaboT -- GPT-4o     & 83.67 & 81.82 (T1) & 89.47 (T2--3) & 78.95 (T2)   \\
\botrule
\end{tabular*}
\footnotetext{T1--T4: error type codes defined in Section~\ref{subsec:errors}.}
\end{table}

\subsection{Error analysis}\label{subsec:errors}

Manual review of failures revealed four primary error types, which informed subsequent prompt refinements:

\begin{itemize}
  \item \textbf{Type 1 -- Validator misclassification}: valid questions incorrectly rejected as out of scope (e.g., legitimate queries about plant metabolites).
  \item \textbf{Type 2 -- Incorrect SPARQL}: queries with incompatible properties or malformed syntax.
  \item \textbf{Type 3 -- Missing KG Agent invocation}: the \textit{Supervisor} failed to call the \textit{KG Agent} for queries requiring entity resolution.
  \item \textbf{Type 4 -- Inappropriate tool selection in the KG Agent}: e.g., using \textit{TaxonResolver} for chemical entities.
\end{itemize}

Type 1 and Type 2 errors dominated for the weaker model (GPT-4o mini), whereas the GPT-4o multi-agent configuration was left almost exclusively with Type 2 errors on the hardest questions.

\subsection{Processing time and operational cost}\label{subsec:cost}

Average query processing time was 10.04~s for the single-shot GPT-4o approach, 36.28~s for the GPT-4o mini multi-agent configuration, and 80.45~s for the GPT-4o multi-agent configuration. Processing time for single-shot GPT-4o was measured from prompt submission until SPARQL generation; for GPT-4o mini, only runs in which all agents were invoked were counted (in many cases the \textit{Validator Agent} halted the workflow before it reached the \textit{SPARQL Query Runner}); for the GPT-4o multi-agent configuration, time was measured for all questions end-to-end.

The MetaboT multi-agent architecture requires multiple LLM calls per query: a typical question incurs five to eight calls depending on complexity. Providing detailed schema information and examples in prompts increases token counts and operational cost. On average, each complete query-processing cycle consumes approximately 22,240 tokens (prompt plus completion), translating to an average cost of \$0.06 per question at the time of evaluation. The maximum observed cost was \$0.092 for a high-complexity question; the minimum was \$0.041 for a low-complexity question.

\subsection{Representative interactions}\label{subsec:interactions}

Figure~\ref{fig:interactions} illustrates two representative MetaboT sessions. We further illustrate the full process with the question \textit{``Which lab extracts have bioassay results with inhibition percentages above 50\% against Leishmania donovani?''} Upon submission, this question is first classified by the \textit{Entry Agent} and validated by the \textit{Validator Agent} to confirm its relevance to the ENPKG schema. The \textit{Supervisor Agent} identifies that the question includes a biological target---\textit{Leishmania donovani}---that must be resolved into a standardised identifier and therefore invokes the \textit{KG Agent}, which calls the \textit{TargetResolver} tool to retrieve the corresponding ChEMBL target IRI. The resolved IRI, together with the original question and schema information, is passed to the \textit{SPARQL Query Runner Agent}, which uses \textit{GraphSparqlQAChain} to construct and execute a SPARQL query against the ENPKG endpoint. The resulting lab extracts meeting the specified criteria are stored in CSV format; when further interpretation is requested, the \textit{Supervisor} calls the \textit{Interpreter Agent}, which opens the CSV and produces a summary or visualisation such as a bar chart. This example highlights the coordinated functioning of agents and tools in transforming a high-level natural-language query into a precise semantic search over the knowledge graph.

%%=============================================================%%
%% Discussion
%%=============================================================%%

\section{Discussion}\label{sec:discussion}

\subsection{Contribution relative to existing software}\label{subsec:comparison}

MetaboT differs from prior single-model and RAG-based approaches to natural-language-to-SPARQL translation~\cite{Rony:2022:SGPT,Meyer:2024:SPARQL-LLM,Emonet:2024:ExpasyGPT,Sima:2023:AI-chatbots-KG} by decomposing the task into specialised agents, each with a single responsibility, and by grounding identifier choices in authoritative external resources (Wikidata, ChEMBL, NPClassifier, the GNPS API) rather than in the LLM's parametric memory. In our evaluation, this architectural separation produces a 10-fold improvement in accuracy over single-shot GPT-4o (83.67\% vs 8.16\%) and preserves accuracy on high-complexity queries that defeat the standalone model (78.95\% vs 0\%).

MetaboT is complementary to the BioChatter~\cite{Lobentanzer:2025:BioChatter} framework for biomedical LLM applications and to the BioCypher~\cite{Lobentanzer:2023:BioCypher} ecosystem for knowledge-graph construction: both target biomedical data at large scale, whereas MetaboT is specifically tuned to the schema, identifiers, and analytical patterns of mass spectrometry metabolomics knowledge graphs. ExpasyGPT~\cite{Emonet:2024:ExpasyGPT} shares the objective of LLM-mediated SPARQL over federated bioinformatics resources; relative to that line of work, MetaboT trades federated coverage for tight integration with a metabolomics KG, explicit entity resolution, and a transparent multi-agent workflow that can be inspected and extended agent-by-agent. Compared with tools that frame chemistry tasks around agentic LLMs (e.g., for synthesis planning), MetaboT addresses the distinct problem of querying rich, schema-constrained semantic graphs rather than invoking external computational chemistry tools. These distinctions, taken together, situate MetaboT as a targeted cheminformatics software contribution for knowledge-graph-based metabolomics.

\subsection{Scalability and adaptability}\label{subsec:scalability}

A critical strength of MetaboT lies in its scalability and adaptability. Although our current demonstration focuses on ENPKG, the modular design supports extension to other mass spectrometry-based knowledge graphs with minimal reconfiguration: agents and tools can be reused, while prompts, schema fragments, and resolver endpoints are adapted to the new graph. The system can also interface with different LLM backends, a useful property given the rapidly evolving LLM landscape~\cite{Chen:2025:LLM-multiagent-survey,Guo:2024:LLM-multiagents-survey}, and the multi-agent architecture allows new agents or tools to be introduced without restructuring the orchestrator.

\subsection{Limitations}\label{subsec:limitations}

Several limitations warrant acknowledgement.

\textbf{Hallucinations.} LLMs occasionally generate incorrect objects or properties, leading to errors in query construction (e.g., confusing \texttt{InChIKey} with \texttt{InChIKey2D}, or applying properties to incompatible classes). These hallucinations can also cause agents to invoke inappropriate tools, such as \textit{TaxonResolver} for queries about biological targets.

\textbf{Non-determinism.} The inherent variability of LLM outputs means that identical queries may yield slightly different SPARQL commands, which occasionally benefits exploration but can also reduce consistency. We mitigate this by combining LLM reasoning with deterministic tools for identifier resolution and by extensive prompt engineering.

\textbf{Model dependence and computational cost.} MetaboT currently requires a high-capacity LLM for reliable performance on complex queries, as illustrated by the substantial accuracy gap between GPT-4o mini and GPT-4o within the same architecture. Each query incurs five to eight LLM calls, an average of $\sim$22,240 tokens, and $\sim$\$0.06 at the time of evaluation---a non-trivial barrier for some research communities~\cite{Sima:2023:AI-chatbots-KG}. Agent prompts were tuned for specific backends and may require adaptation when switching models. Cost and robustness could be improved with hybrid approaches combining several models of varying capacity alongside automated prompt optimisation.

\textbf{Single-graph querying.} MetaboT is currently restricted to querying one knowledge graph at a time. Researchers must manually connect ENPKG data to external resources such as ChEMBL~\cite{Zdrazil:2024:ChEMBL}, Wikidata~\cite{Waagmeester:2020:Wikidata}, or LOTUS~\cite{Rutz:2022:LOTUS} when federated answers are needed, because the system does not yet generate federated SPARQL queries. While critical external information has been integrated directly into ENPKG, implementing true federated-query capabilities would require efficient KG indexing, graph summaries for source selection, and sophisticated query-planning algorithms~\cite{Heling:2021:federated-SPARQL}. A dedicated federated-query module could then emit \texttt{SERVICE} clauses following the SPARQL 1.2 Federated Query specification (\url{https://www.w3.org/TR/sparql12-federated-query/}) to bridge ENPKG with complementary resources such as LOTUS.

\textbf{Evaluation scope.} The evaluation focuses on SPARQL query generation and does not benchmark the \textit{Interpreter Agent}'s summaries or visualisations. Downstream interpretation accuracy is left to future work.

\subsection{Future directions}\label{subsec:future}

Looking ahead, we envision MetaboT evolving into a broader toolbox for mass spectrometry data analysis, integrating agents and components dedicated to commonly used computational metabolomics approaches. Future developments will focus on automated benchmarking for continuous performance assessment, expanded interoperability with other metabolomics tools and knowledge graphs, and integration of emerging AI capabilities for cheminformatics and biomedical hypothesis generation~\cite{Lobentanzer:2025:BioChatter}. These extensions will be pursued within the MetaboLinkAI programme, which brings together eight research institutions aiming to transform metabolomics data analysis through AI and knowledge-graph technologies.

%%=============================================================%%
%% Conclusions
%%=============================================================%%

\section{Conclusions}\label{sec:conclusions}

MetaboT is a reproducible multi-agent LLM framework that bridges the gap between specialised SPARQL query writing and intuitive conversational interactions with mass spectrometry metabolomics knowledge graphs. By combining orchestrated entity resolution with schema-aware SPARQL generation and an iterative refinement loop, a multi-agent architecture powered by a high-capacity reasoning model achieves 83.67\% accuracy in translating natural-language questions into executable queries on ENPKG---a 10-fold improvement over a single-shot baseline on the same benchmark. Released as open-source software with a public web demonstrator, a tagged evaluated version, and a reusable benchmark, MetaboT lowers the technical barrier for domain scientists, enables systematic semantic data mining without specialised programming expertise, and provides a concrete, extensible proof of concept for the broader MetaboLinkAI programme of AI-assisted metabolomics.

%%=============================================================%%
%% Availability and requirements
%%=============================================================%%

\section*{Availability and requirements}

\begin{itemize}
  \item \textbf{Project name:} MetaboT
  \item \textbf{Project home page:} \url{https://github.com/HolobiomicsLab/MetaboT}
  \item \textbf{Public demonstrator:} \url{https://metabot.holobiomicslab.eu/}
  \item \textbf{Archived evaluated version:} Zenodo, DOI 10.5281/zenodo.19715403 (tag \texttt{v1.0-evaluation})
  \item \textbf{Operating systems tested:} macOS Sonoma (14.5), Ubuntu 22.04 LTS, Debian 11
  \item \textbf{Programming language:} Python (3.10)
  \item \textbf{Other requirements:} LangChain 0.3.27, LangGraph 0.3.34, FAISS 1.8.0, OpenAI Python library 2.32.0, pandas 2.2.1, Plotly 5.20.0. Hardware: modern CPU (Intel Core i7 13th generation or equivalent); minimum 8~GB RAM. Access to an LLM API endpoint (OpenAI GPT-5.5 recommended, minimum GPT-4o) is required for full functionality; a demonstration mode with canned responses is provided for local development without API access. A Docker configuration is included for containerised deployment.
  \item \textbf{Access to ENPKG:} \url{https://enpkg.commons-lab.org/graphdb/repositories/ENPKG}
  \item \textbf{License:} Apache 2.0
  \item \textbf{Any restrictions to use by non-academics:} none beyond the Apache 2.0 license.
  \item \textbf{Reviewer access:} during peer review, a reviewer-anonymous contributor key is provided for the public web demonstrator; please see the cover letter for the current key.
\end{itemize}

%%=============================================================%%
%% Backmatter
%%=============================================================%%

\backmatter

\bmhead{Supplementary information}

\textbf{Additional file 1 (PDF):} Supplementary material. Figure~S1---Detailed agent-interaction workflow with decision points, including the Entry $\to$ Validator $\to$ Supervisor routing, entity-resolution branch via the KG Agent, SPARQL query generation and refinement, and interpretation by the Interpreter Agent.
\textbf{Additional file 2 (CSV):} Evaluation dataset of 50 natural-language questions with reference SPARQL queries and complexity categorisation (also archived at \url{https://doi.org/10.5281/zenodo.19715403}).

\bmhead{Acknowledgements}

The figures were composed using Microsoft PowerPoint and include components adapted from BioRender (\url{https://www.biorender.com}), icons sourced from Freepik (\url{https://www.freepik.com}), icons available in the PowerPoint standard library, and diagrams generated using Mermaid (\url{https://mermaid.js.org}).

\section*{Declarations}

\subsection*{Ethics approval and consent to participate}

Not applicable.

\subsection*{Consent for publication}

Not applicable.

\subsection*{Availability of data and materials}

All data and software necessary to reproduce the results are publicly available. The Experimental Natural Products Knowledge Graph (ENPKG), which integrates mass spectrometry data with taxonomic information, chemical structures, and biological activities of natural products derived from 1,600 plant extracts, is accessible through the public SPARQL endpoint at \url{https://enpkg.commons-lab.org/graphdb/repositories/ENPKG} and was originally published by Gaudry et al.~\cite{Gaudry:2024:ENPKG}. The synthetic evaluation dataset of 50 reference questions with corresponding SPARQL queries, categorised by complexity, is released in the project repository at \url{https://github.com/HolobiomicsLab/MetaboT} (\texttt{app/data/evaluation\_dataset.csv}) and archived on Zenodo at \url{https://doi.org/10.5281/zenodo.19715403}. The Zenodo archive additionally includes: (i) \texttt{Big\_Benchmark\_traces\_13.11.2024\_gpt\_4o.csv}, comprising outputs from the LangSmith automated evaluation of the MetaboT system using GPT-4o as the LLM for all agents, including generated answers and LLM-based evaluation scores; (ii) \texttt{Big\_Benchmark\_traces\_06.11.2024\_gpt4o\_mini.csv}, comprising analogous LangSmith evaluation outputs using GPT-4o mini; (iii) \texttt{manual\_evaluation\_metabot\_gpt\_4o.csv}, containing manual evaluation results for MetaboT with GPT-4o, where incorrect cases are explicitly marked in the Comments column, and the Issue/error column details the nature of the failure (e.g., incorrect SPARQL query generation or missing agent calls); (iv) \texttt{manual\_evaluation\_baseline\_gpt4o\_alone.csv}, containing SPARQL queries generated by GPT-4o alone (without MetaboT), with correctness indicated in the Comments column; and (v) \texttt{manual\_evaluation\_metabot\_gpt4o\_mini.csv}, containing manual evaluation results for MetaboT with GPT-4o mini, with incorrect cases annotated as described above. Taxonomic information is retrieved from Wikidata and biological-target information from ChEMBL; both are open-access resources. Researchers who wish to apply MetaboT to their own data can follow the conversion pipeline described in the ENPKG documentation (\url{https://github.com/enpkg/enpkg_full}) to transform mass spectrometry processing and annotation results into a compatible knowledge-graph format.

The complete source code for MetaboT is available as an open-source project under the Apache 2.0 License at \url{https://github.com/HolobiomicsLab/MetaboT}. The repository contains all component code, configuration files, agent implementations, prompt templates, and evaluation scripts necessary to reproduce the system described in this article. It also includes installation instructions, Docker configuration files, and Conda environment files that specify exact version requirements. The version of the code used for the analyses and evaluation presented here is tagged \texttt{v1.0-evaluation} and archived on Zenodo (DOI 10.5281/zenodo.19715403). The evaluation procedure is executed via the script at \url{https://github.com/HolobiomicsLab/MetaboT/blob/main/app/core/tests/evaluation.py}. A public instance of MetaboT connected to ENPKG is accessible at \url{https://metabot.holobiomicslab.eu/} and allows researchers to test the system without local installation.

\subsection*{Competing interests}

F.G. is co-founder and co-scientific advisor of :probabl.; E.T. is an employee of :probabl. All other authors declare no competing interests.

\subsection*{Funding}

This work has been supported by the French government through the UCAJEDI Investments in the Future project managed by the National Research Agency (ANR) with reference number ANR-15-IDEX-01, through the France 2030 investment plan via the MetaboLinkAI bilateral project (ANR-24-CE93-0012-01 and SNSF 10002786), and through the 3IA C\^ote d'Azur programme (ANR-23-IACL-0001). L.P. was funded by the Luxembourg National Research Fund (FNR; project 17994255). E.T. received funding from the MIT International Science and Technology Initiatives through the MIT-France Program. O.K., F.M., M.P., J.-L.W. and L.-F.N. were supported by the Swiss National Science Foundation (SNSF project CRSII5\_189921/1). W.B. was supported by the Research Foundation--Flanders (FWO G087625N) and the University of Antwerp Research Fund. The funders had no role in the study design, data collection and analysis, decision to publish, or manuscript preparation.

\subsection*{Authors' contributions}

M.B., F.G., and L.-F.N. conceptualised the method. S.H., W.B., J.-L.W., and M.P. contributed to method conception and evaluation. M.B., L.P., B.N., E.T., F.Mi., and M.F. implemented the software. M.B. performed the analysis. M.B., Y.T., Y.Z.C., O.K., F.Me., M.L., and T.J. tested the method. W.B., F.G., and L.-F.N. co-supervised the project. M.B., W.B., F.G., and L.-F.N. wrote and edited the manuscript. All authors read and approved the final manuscript.

\subsection*{Use of large language models}

LLMs (OpenAI GPT-5.5 and GPT-5.4 mini) are components of the software system described in this article; their roles and versions are documented in the \textit{Implementation} and \textit{Results} sections. No LLMs were used to generate scientific content, results, or analyses beyond their described role as system components. LLMs were used only for light editorial assistance (grammar and wording) during manuscript preparation; all content, conclusions, and interpretations are the authors' own.

\subsection*{Authors' information}

Not applicable.

%%=============================================================%%
%% Bibliography
%%=============================================================%%

\bibliography{MetaboT}

\end{document}